\documentclass[conference]{IEEEtran}
\usepackage{cite}
\usepackage{amsmath,amssymb,amsfonts}
\usepackage{graphicx}
\usepackage{textcomp}
\usepackage{xcolor}
\usepackage{algorithm}
\usepackage{algpseudocode}
\usepackage{booktabs}

\def\BibTeX{{\rm B\kern-.05em{\sc i\kern-.025em b}\kern-.08em
    T\kern-.1667em\lower.7ex\hbox{E}\kern-.125emX}}

\begin{document}
\IEEEoverridecommandlockouts
% \IEEEpubid{\makebox[\columnwidth]{ 979-8-3195-0489-0/26/\$31.00~\copyright~2026 IEEE \hfill} \hspace{\columnsep}\makebox[\columnwidth]{ }}
\title{MoE Router-Guided Clustering for Heterogeneous Federated Instruction Tuning}
\author{ Ankita Sharma$^{1}$, Bahar Farahani$^{2}$, Sanaz Rahimi Moosavi$^{3}$, Amir Rahmani$^{4}$, \\ Farshad Firouzi$^{1,5}$, and Krishnendu Chakrabarty$^{1}$ \\[1ex] \small $^{1}$Arizona State University \quad $^{2}$Member, IEEE \quad $^{3}$California State University, Dominguez Hills \\ \quad $^{4}$University of California, Irvine \quad $^{5}$Johns Hopkins University}
\maketitle
\IEEEpubidadjcol
\begin{abstract}
% Say data-driven signal as in , activation patterns that we use to make clusters, why client-based and why expert-based.
Federated instruction fine-tuning enables Large Language Models (LLMs) to adapt to decentralized, privacy-sensitive data without requiring data sharing. Recent Mixture-of-Experts (MoE) LLMs are particularly attractive for federated learning because their sparse activation reduces computation and communication while scaling model capacity. However, existing federated MoE methods primarily focus on parameter aggregation and personalization, overlooking the routing behavior of MoE models as a source of information for client collaboration. Under heterogeneous instruction distributions, indiscriminate aggregation can lead to negative transfer, highlighting the need to identify which clients should collaborate during federated optimization. We propose \textit{ClientMorpher}, a routing-aware, personalized federated instruction fine-tuning framework that leverages routing signatures from pretrained MoE models to organize client collaboration prior to aggregation. We investigate two complementary clustering strategies: \textit{ClientMorpher-C}, which directly clusters clients using expert activation profiles, and \textit{ClientMorpher-E}, which first clusters experts based on their cross-client usage signatures and then derives client collaboration groups. Together, these strategies exploit routing behavior as a structural signal for personalization while maintaining the communication efficiency of sparse MoE fine-tuning. We evaluate \textit{ClientMorpher} for federated instruction fine-tuning on the Databricks Dolly-15K dataset, using pathological and Dirichlet-based heterogeneous client distributions across multiple instruction-following tasks. Experimental results demonstrate that routing-aware collaboration consistently improves personalized performance compared to conventional federated averaging and local training, while maintaining the same communication cost. Furthermore, our study shows that client-centric and expert-centric clustering capture complementary forms of task similarity, providing an effective and scalable approach for personalized federated instruction fine-tuning of sparse MoE LLMs.
\end{abstract}


\section{Introduction}
\label{sec:introduction}
%LLM → Federated Instruction Tuning → MoE Routing → Why FedAvg fails → Why routing is useful → Gap → ClientMorpher
Large Language Models (LLMs) have significantly advanced natural language processing by achieving state-of-the-art performance across a wide range of instruction-following tasks, including text classification, summarization, question answering, and information extraction. As these models become increasingly integrated into real-world applications, fine-tuning them on organization-specific data has become essential for adapting to specialized domains and user requirements~\cite{firouzi2026generative}. However, such data are often distributed across multiple organizations, institutions, or edge devices and cannot be centralized because of privacy regulations, proprietary constraints, or communication limitations. Federated Learning (FL)~\cite{mcmahan2017communication,jiang2024federated,jiang2023low} addresses this challenge by enabling multiple clients to collaboratively fine-tune a shared model while keeping local data decentralized. More recently, Mixture-of-Experts (MoE) architectures~\cite{shazeer2017outrageously} have emerged as an attractive foundation for large-scale federated deployment because they activate only a sparse subset of experts for each input token, allowing models to scale to billions of parameters while maintaining efficient computation and communication~\cite{fedus2022switch}. Beyond computational efficiency, the routing mechanism of MoE models determines which experts process each token, implicitly capturing characteristics of the underlying instruction distribution. This routing behavior provides an additional source of information that is unique to sparse MoE models~\cite{wang2026myth,avinash2026task} and has not yet been fully exploited in federated instruction fine-tuning.
\begin{figure}[!]
    \centering
    \includegraphics[width=8.5cm]{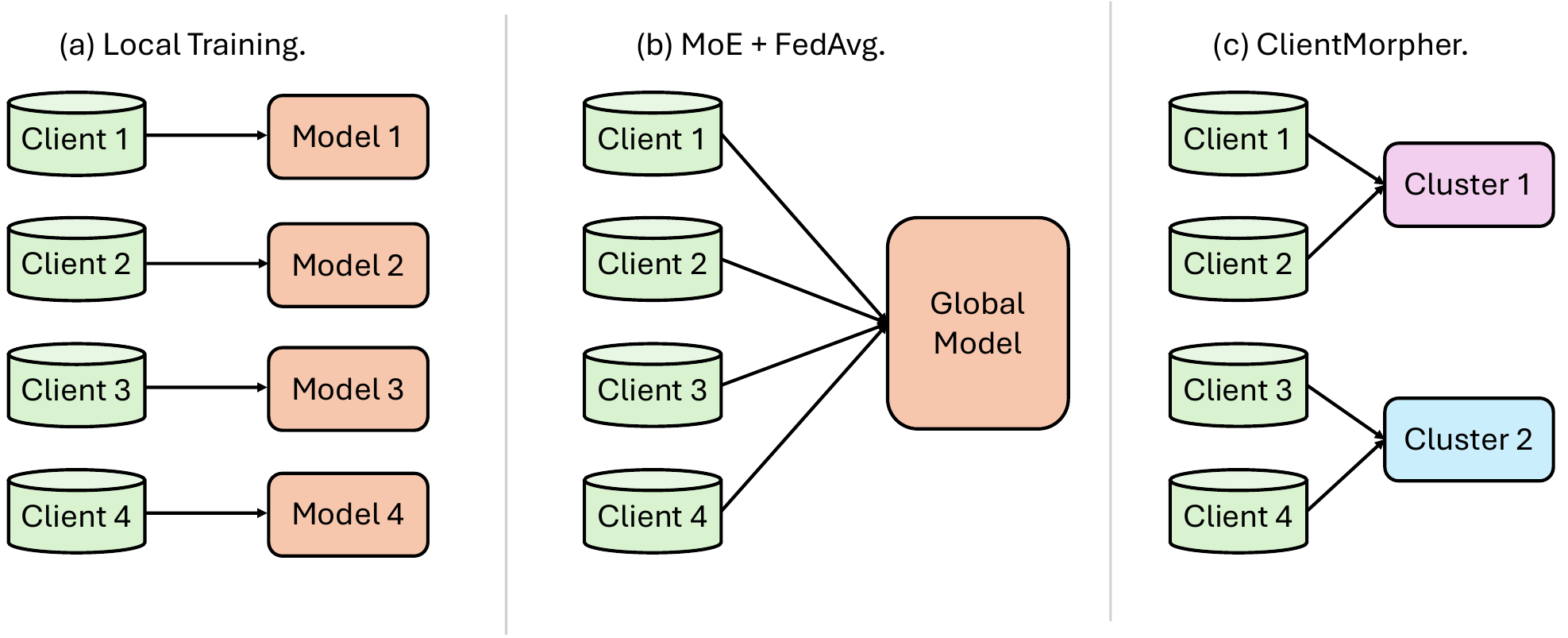}
    \caption{Motivation for ClientMorpher. (a) Local training provides personalization but no collaboration. (b) FedAvg aggregates all MoE experts across clients, ignoring expert usage patterns. (c) ClientMorpher clusters clients using sparse expert activation profiles and performs cluster-wise aggregation of shared parameters.}
    \label{fig:motivation}
\end{figure}

\begin{figure*}[t]
    \centering
    \includegraphics[width=\textwidth]{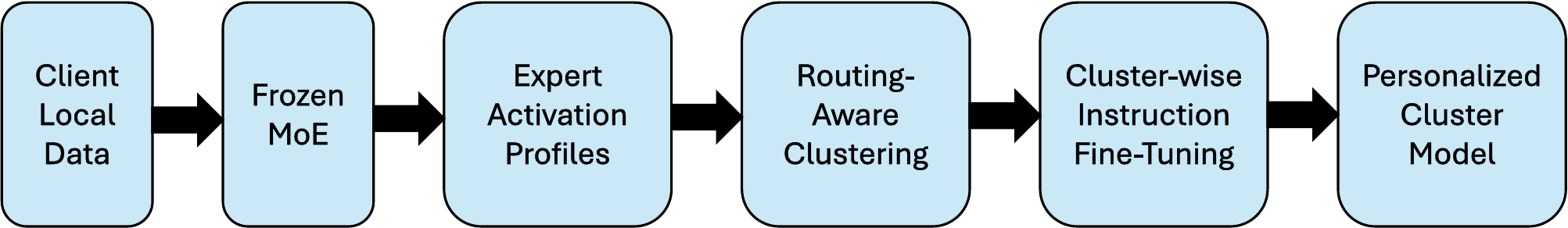}
    \caption{The end-to-end workflow of the ClientMorpher Framework.}
    \label{fig:framework_overview}
\end{figure*}

Despite these advantages, federated instruction fine-tuning presents significant challenges under heterogeneous client distributions. In practical deployments, clients rarely possess identically distributed data; instead, they often specialize in different domains or instruction types, such as summarization, classification, question answering, or information extraction, as depicted in Fig.~\ref{fig:motivation}. Conventional Federated Averaging (FedAvg) aggregates updates from all participating clients into a single global model, implicitly assuming that every client should contribute equally to a shared representation. Under highly heterogeneous settings, however, indiscriminate aggregation frequently introduces negative transfer, reducing both personalization and overall model quality~\cite{kairouz2021advances,li2020federated}. Recent personalized federated learning methods~\cite{tan2022towards} attempt to alleviate this issue by adapting subsets of parameters or maintaining client-specific models, while federated MoE frameworks such as FedMoE~\cite{mei2024fedmoe}, FedMoE-DA~\cite{zhan2024fedmoe}, and OpenFedLLM~\cite{ye2024openfedllm} further exploit sparse architectures to reduce communication overhead and improve scalability. Nevertheless, these methods largely treat collaboration as a predefined optimization process and overlook  MoE model's routing behavior. In particular, existing approaches do not investigate whether routing patterns can reveal latent similarities among clients before federated optimization begins, nor whether such information can guide more effective collaboration among heterogeneous participants supporting clustering~\cite {ghosh2020efficient}.

Since the router dynamically selects experts for each instruction, clients with similar instruction distributions are expected to exhibit similar routing behavior. These routing signatures, therefore, offer an observable representation of client similarity before federated optimization begins. Based on this insight, we propose ClientMorpher, a routing-aware personalized federated learning framework that leverages routing signatures to form collaboration groups prior to model aggregation. We investigate two complementary perspectives: ClientMorpher-C, which groups clients directly according to their expert activation profiles, and ClientMorpher-E, which first discovers expert communities from expert usage signatures and then derives client collaboration groups. %This formulation enables us to study whether routing-aware collaboration improves personalization under heterogeneous instruction distributions and whether client-centric and expert-centric clustering capture complementary forms of task similarity.

In summary, this paper makes the following contributions:
\begin{itemize}
    \item We propose ClientMorpher, a routing-aware personalized federated instruction fine-tuning framework as illustrated in Fig.~\ref{fig:framework_overview} that exploits the routing behavior of sparse MoE LLMs to guide client collaboration.
    \item We develop two complementary clustering strategies: ClientMorpher-C, which clusters clients using expert activation profiles, and ClientMorpher-E, which clusters experts using cross-client usage signatures before assigning clients to expert-induced collaboration groups.
    \item We conduct an empirical study on heterogeneous splits of the Dolly-15k dataset~\cite{DatabricksBlog2023DollyV2}, comparing local training, MoE-FedAvg, ClientMorpher-C, and ClientMorpher-E across pathological and Dirichlet partitions with varying instruction-task skew.
\end{itemize}
The remainder of this paper is organized as follows. Section~\ref{related-work} reviews related work. Section~\ref{method} presents the proposed framework. Section~\ref{design} describes the experimental setup, evaluation, and results. Finally, Section~\ref{conclusion} concludes the paper.
\section{Related Work}
\label{related-work}
\subsection{Clustering Clients for Personalization}

Client clustering is an effective personalization strategy for federated learning in the presence of statistical heterogeneity. Instead of optimizing a single global model, clustered federated learning groups clients with similar data distributions or optimization behavior and trains a shared model for each cluster, improving convergence while reducing negative transfer. Representative approaches include IFCA~\cite{ghosh2020efficient}, CFL~\cite{Sattler2019ClusteredFL}, FedSEM~\cite{zhou2025fedsem}, FedGroup~\cite{duan2021fedgroup}, and StoCFL~\cite{zeng2025stocfl}, which construct client clusters using gradients, model parameters, feature representations, or optimization dynamics. Although these methods have demonstrated strong performance with conventional neural networks, they are not designed for Mixture-of-Experts (MoE) models, in which sparse expert routing naturally captures functional specialization. However, existing clustering methods fail to leverage routing behavior as a lightweight, semantically meaningful signal to identify similar clients.

\subsection{Federated Learning with Mixture-of-Experts Models}
Mixture-of-Experts (MoE) architectures have recently been introduced into federated learning to increase model capacity while maintaining communication and computational efficiency through sparse expert activation. Existing methods, including PFL-MoE~\cite{guo2020pfl}, FedMoE~\cite{mei2024fedmoe}, Fed-MoE, pFedMoE~\cite{Yi2024pFedMoEDP}, PM-MoE~\cite{jiang2025heterogeneous}, and FLEx~\cite{liu2025flex}, primarily focus on personalized adaptation, expert selection, modular aggregation, or communication-efficient training. While these approaches leverage the MoE routing mechanism during optimization, they largely treat routing as a means of personalization rather than as a source of structural information about client relationships. In contrast, our work utilizes expert-routing statistics to construct expert signatures, discover client similarity through expert-overlap clustering, and perform cluster-personalized federated optimization without exchanging expert parameters.

\section{Methodology}
\label{method}

\begin{figure}[!]
    \centering
    \includegraphics[width=8.5cm]{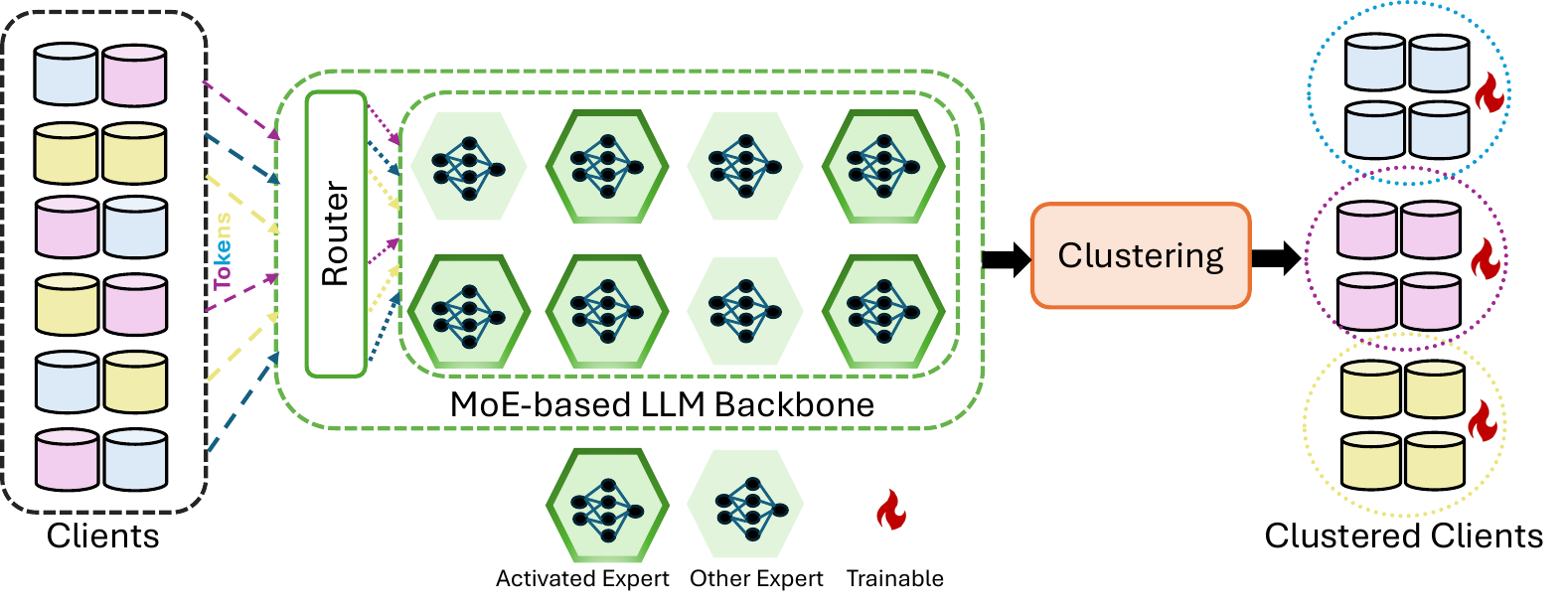}
    \caption{The depiction of router-guided client clustering for MoE-LLMs.}
    \label{fig:framework}
\end{figure}

We present \textit{ClientMorpher}, a routing-aware personalization framework for federated fine-tuning of MoE-LLMs. The core idea is that the router of a pretrained MoE model provides a compact signature of each client's data distribution. Clients who activate similar experts are likely to require similar adaptation behavior, whereas clients who route to different experts should not necessarily share a single globally averaged model. ClientMorpher, therefore, replaces global aggregation with cluster-specific federated optimization as illustrated in Fig.~\ref{fig:framework}.
%This is the client-profile method reported as \texttt{client\_clustering} in the experiments. Second is the expert-overlap method reported as \texttt{expert\_clustering}.
We study two variants of the same framework. \textbf{ClientMorpher-C} clusters clients directly using their expert-activation profiles. \textbf{ClientMorpher-E} first clusters experts according to their cross-client usage signatures and then assigns clients to expert clusters. Both variants share the same profiling step and the same cluster-personalized federated training step; they differ only in how client clusters are constructed, as explained in Algorithm~\ref{alg:clientmorpher_unified}.

\subsection{Federated MoE Setup}
\label{subsec:method_setup}
% In the experiments, $\theta_a$ corresponds to LoRA adapters applied to the non-expert attention projections.
Consider $N$ clients indexed by $i \in \{1,\ldots,N\}$, where client $i$ owns a local dataset $\mathcal{D}_i$. The server initializes a pretrained MoE language model with $E$ routed experts. For an input token, the MoE router produces logits over the experts and selects the top-$k$ experts. The pretrained MoE backbone and expert parameters are kept frozen, and only a small set of adaptation parameters $\theta_a$ is optimized. The objective is to learn personalized adaptation parameters without requiring each client to train in isolation or forcing all clients into a single global FedAvg model. ClientMorpher solves this by learning a cluster assignment $a_i \in \{1,\ldots,K\}$ for each client and maintaining one adaptation state $\theta_a^{(k)}$ per cluster.

\begin{algorithm}[t]
\caption{ The ClientMorpher Workflow.}
\label{alg:clientmorpher_unified}
\footnotesize
\begin{algorithmic}[1]
\Require Client datasets $\{\mathcal{D}_i\}_{i=1}^{N}$; pretrained MoE model with $E$ experts; top-$k$ router selections; profile budget $B$; clusters $K$; rounds $R$; local steps $L$; variant $v\in\{\mathrm{C},\mathrm{E}\}$
\Ensure Cluster-specific adaptation states $\{\theta_a^{(k,R)}\}_{k=1}^{K}$
\State Freeze pretrained MoE backbone and expert parameters
\State Initialize trainable adaptation parameters $\theta_a$
\Comment{Shared routing profiling}
\For{each client $i \in \{1,\ldots,N\}$}
    \State Run up to $B$ batches from $\mathcal{D}_i$ through the frozen MoE router
    \State Count expert selections $c_{i,e}=\sum_{t\in\mathcal{T}_i}\mathbf{1}[e\in\mathrm{TopK}_{i,t}]$
    \State Normalize $u_{i,e}=c_{i,e}/\sum_{e'=1}^{E}c_{i,e'}$
    \State Store client profile $\mathbf{u}_i=[u_{i,1},\ldots,u_{i,E}]$
\EndFor
\State Build client-expert matrix $\mathbf{U}\in[0,1]^{N\times E}$ from rows $\mathbf{u}_i$
\Comment{Variant-specific clustering}
\If{$v=\mathrm{C}$}
    \State Apply K-Means to client profiles $\{\mathbf{u}_i\}_{i=1}^{N}$
    \State Assign clients directly:
    $a_i=\arg\min_{\ell\in\{1,\ldots,K\}}\|\mathbf{u}_i-\boldsymbol{\mu}_{\ell}\|_2^2$
\Else
    \For{each expert $e \in \{1,\ldots,E\}$}
        \State Construct and normalize expert signature $\mathbf{s}_e=\mathbf{U}_{:,e}/\|\mathbf{U}_{:,e}\|_2$
    \EndFor
    \State Apply K-Means to expert signatures $\{\mathbf{s}_e\}_{e=1}^{E}$
    \State Obtain expert labels $z_e=\arg\min_{\ell\in\{1,\ldots,K\}}\|\mathbf{s}_e-\boldsymbol{\nu}_{\ell}\|_2^2$
    \State Build $\mathbf{M}\in\{0,1\}^{E\times K}$ where $M_{e,\ell}=1$ iff $z_e=\ell$
    \State Assign clients by expert-cluster mass:
    $a_i=\arg\max_{\ell\in\{1,\ldots,K\}}\sum_{e=1}^{E}u_{i,e}M_{e,\ell}$
\EndIf
\State Form client clusters $\mathcal{C}_k=\{i:a_i=k\}$
\Comment{Shared cluster-PFL}
\For{each cluster $k \in \{1,\ldots,K\}$}
    \State Initialize cluster state $\theta_a^{(k,0)}\leftarrow\theta_a$
\EndFor
\For{round $r=0,\ldots,R-1$}
    \For{each non-empty cluster $\mathcal{C}_k$}
        \For{each client $i\in\mathcal{C}_k$}
            \State Send $\theta_a^{(k,r)}$ to client $i$
            \State $\theta_{a,i}^{(r+1)}\leftarrow\mathrm{LocalTrain}(\theta_a^{(k,r)},\mathcal{D}_i,L)$
        \EndFor
        \State Aggregate within cluster:
        $\theta_a^{(k,r+1)}\leftarrow\sum_{i\in\mathcal{C}_k}\frac{n_i}{\sum_{j\in\mathcal{C}_k}n_j}\theta_{a,i}^{(r+1)}$
    \EndFor
\EndFor
\State \Return $\{\theta_a^{(k,R)}\}_{k=1}^{K}$
\end{algorithmic}
\end{algorithm}

\subsection{Phase 0: Shared Client Expert-Usage Profiling}
\label{subsec:phase0}

Both ClientMorpher-C and ClientMorpher-E begin by profiling how each client's
local data activates the frozen MoE router. For client $i$, we run up to $B$
batches from $\mathcal{D}_i$ through the model without updating parameters. Let
$\mathcal{T}_{i}$ denote the profiled token positions across all captured MoE
layers. For token position $t \in \mathcal{T}_{i}$, let
$\mathrm{TopK}_{i,t} \subseteq \{1,\ldots,E\}$ be the set of top-$k$ experts
selected by the router.

We count how often each expert is selected:
\begin{equation}
c_{i,e}
= \sum_{t \in \mathcal{T}_{i}}
\mathbf{1}\left[e \in \mathrm{TopK}_{i,t}\right],
\end{equation}
where $c_{i,e}$ is the number of routing assignments from client $i$ to expert
$e$. The normalized expert-usage profile for client $i$ is
\begin{equation}
u_{i,e}
=
\frac{c_{i,e}}{\sum_{e'=1}^{E} c_{i,e'}},
\qquad
\mathbf{u}_i = [u_{i,1},\ldots,u_{i,E}] \in [0,1]^E.
\end{equation}
The vectors are stacked into the client-expert usage matrix
\begin{equation}
\mathbf{U}
=
\begin{bmatrix}
\mathbf{u}_1^\top \\
\mathbf{u}_2^\top \\
\vdots \\
\mathbf{u}_N^\top
\end{bmatrix}
\in [0,1]^{N \times E}.
\end{equation}
The $i$-th row of $\mathbf{U}$ represents the routing profile of client $i$.
This matrix is the common input to both ClientMorpher variants.

\subsection{ClientMorpher-C: Direct Client-Profile Clustering}
\label{subsec:clientmorpher_c}

ClientMorpher-C directly clusters clients in the expert-usage space. Each
client is represented by its row vector $\mathbf{u}_i$ from the usage matrix
$\mathbf{U}$. We apply K-Means to the set of client profiles $\{\mathbf{u}_i\}_{i=1}^{N}$:
\begin{equation}
a_i
=
\arg\min_{k \in \{1,\ldots,K\}}
\left\| \mathbf{u}_i - \boldsymbol{\mu}_k \right\|_2^2,
\end{equation}
where $\boldsymbol{\mu}_k$ is the centroid of cluster $k$. This produces client clusters
$\mathcal{C}_k = \{i : a_i = k\}$. This variant treats the entire expert-activation profile as a client signature. If two clients route to experts with similar frequencies, they are assigned to the same aggregation group. %In the experimental section, this method is denoted \textbf{ClientMorpher-C}. %It is the main ClientMorpher variant corresponding to the historical \texttt{clientmorpher} results and the current \texttt{client\_clustering} mode.

\subsection{ClientMorpher-E: Expert-Overlap Clustering}
\label{subsec:clientmorpher_e}

ClientMorpher-E uses the same usage matrix $\mathbf{U}$, but it clusters experts before assigning clients. The routing signature of expert $e$ is the cross-client usage vector given by the $e$-th column of $\mathbf{U}$:
$\mathbf{s}_e = \mathbf{U}_{:,e} \in \mathbb{R}^{N}.$ Each expert signature is normalized before clustering. We then compute pairwise expert similarity using cosine similarity. The experts are clustered into $K$ expert groups. We use K-Means on the expert signatures:
% \begin{equation}
% S_{e,e'}
% =
% \frac{\mathbf{s}_e^\top \mathbf{s}_{e'}}
% {\|\mathbf{s}_e\|_2 \|\mathbf{s}_{e'}\|_2}.
% \end{equation}
%where large values indicate that two experts serve similar sets of clients.

\begin{equation}
z_e
=
\arg\min_{k \in \{1,\ldots,K\}}
\left\| \mathbf{s}_e - \boldsymbol{\nu}_k \right\|_2^2,
\end{equation}
where $z_e$ is the expert-cluster assignment and $\boldsymbol{\nu}_k$ is an expert-cluster centroid. We define an expert-membership matrix $\mathbf{M}\in\{0,1\}^{E \times K}$:
\begin{equation}
M_{e,k} =
\begin{cases}
1, & z_e = k, \\
0, & \text{otherwise}.
\end{cases}
\end{equation}

Each client is then assigned to the expert cluster that receives the largest mass from its local routing profile. The score of client $i$ for the expert cluster
$k$ is
\begin{equation}
q_{i,k}
=
\sum_{e=1}^{E} u_{i,e} M_{e,k}.
\end{equation}
The final client assignment is
$a_i=\arg\max_{k \in \{1,\ldots,K\}} q_{i,k}.$ The resulting client clusters are again $\mathcal{C}_k = \{i : a_i = k\}$. This variant is denoted \textbf{ClientMorpher-E} in the experimental section. Unlike ClientMorpher-C, which clusters clients directly, ClientMorpher-E first discovers groups of experts that are shared across clients and then derives client groups from those expert clusters. %This makes it an expert-overlap variant of ClientMorpher rather than a separate baseline.

\subsection{Shared Phase 2: Cluster-Personalized Federated Learning}
\label{subsec:phase2}

After either ClientMorpher-C or ClientMorpher-E produces client clusters, the training phase is identical. The server maintains one adaptation state $\theta_a^{(k)}$ for each non-empty cluster $\mathcal{C}_k$. At communication round $r$, the server sends $\theta_a^{(k,r)}$ to clients in cluster $\mathcal{C}_k$. Each client performs local optimization on its dataset $\mathcal{D}_i$ for a fixed number of local steps:
\begin{equation}
\theta_{a,i}^{(r+1)}
=
\mathrm{LocalTrain}
\left(\theta_a^{(a_i,r)}, \mathcal{D}_i\right).
\end{equation}
The server then aggregates updates only within the same cluster using FedAvg:
\begin{equation}
\theta_a^{(k,r+1)}
=
\sum_{i \in \mathcal{C}_k}
\frac{n_i}{\sum_{j \in \mathcal{C}_k} n_j}
\theta_{a,i}^{(r+1)},
\end{equation}
where $n_i = |\mathcal{D}_i|$ is the number of local training samples for client $i$. If a cluster contains a single client, its cluster state is simply that client's locally trained state.

\begin{table}[t]
\centering
\caption{Comparison of the two clustering strategies.}
\label{tab:cc_vs_ec}
\scriptsize
\setlength{\tabcolsep}{3pt}
\renewcommand{\arraystretch}{1.1}
\begin{tabular}{lcc}
\toprule
 & CC & EC \\
\midrule
K-Means input & Client profiles & Expert signatures \\
Input shape & $N\times E$ & $E\times N$ \\
Output & Client labels $a_i$ & Expert labels $z_e$ \\
Assignment & Direct ($a_i$) &
$\arg\max_k \sum_e u_{i,e}M_{e,k}$ \\
Aggregation & Cluster-wise FedAvg & Cluster-wise FedAvg \\
\bottomrule
\end{tabular}
\end{table}

\section{Experimental Design and Evaluation Results}
\label{design}
%results/cm_pathological_10r_fast_final; /home/ashar236/Projects/Coins/clientmorpher/results/cm_pathological_10r_fast_final/summary_all_experiments.json
% clustering C and E cn have more columns about clustering quality, like nmi, ari, purity, intra and inter similarity
\begin{table}[t]
\centering
\caption{4-client pathological split. All federated methods exchange identical LoRA parameters (0.0878\% of model/round).}
\label{tab:baseline_results}
\scriptsize
\resizebox{\columnwidth}{!}{%
\begin{tabular}{lcccccc}
\toprule
Method & CLF & CQA & IE & SUM & Avg. & Comm.\% \\
% \midrule
% FedAvg (full model) & -- & -- & -- & -- & -- & -- & 100.0 \\
% FedAvg (dense non-expert) & -- & -- & -- & -- & -- & -- & 13.06 \\
%method_cost_pct
\midrule
Local & 0.6036 & 0.4096 & 0.5087 & 0.4075 & 0.4824 & 0.00 \\
MoE-FedAvg (LoRA) & 0.6299 & 0.4232 & 0.5085 & 0.4090 & 0.4927 &  0.0878 \\
ClientMorpher-C & 0.5937 & 0.3785 & \textbf{0.5243} & 0.3960 & 0.4731 & 0.0878 \\
ClientMorpher-E & \textbf{0.6411} & \textbf{0.4254} & 0.5099 & \textbf{0.4181} & \textbf{0.4986} & 0.0878 \\
\bottomrule
\end{tabular}}
\end{table}
%"All federated methods exchange the same LoRA parameters. Therefore, the performance improvements of ClientMorpher arise from better collaboration rather than additional communication."

%/home/ashar236/Projects/Coins/clientmorpher/outputs_old/dolly_dirichlet/split=dirichlet/mode=expert_clustering/clients=8/clusters=4/alpha=0.1/seed=42/topk=4/lora_rank=32/lr=4e-05/rounds=100/metrics.json 
%/home/ashar236/Projects/Coins/clientmorpher/outputs_old/dolly_dirichlet/split=dirichlet/mode=expert_clustering/clients=8/clusters=4/alpha=1.0/seed=42/topk=4/lora_rank=32/lr=4e-05/rounds=100/metrics.json use round 30 value for comparision
% alpha =0.1 /home/ashar236/Projects/Coins/clientmorpher/outputs_old/dolly_dirichlet_client/split=dirichlet/mode=client_clustering/clients=8/clusters=4/alpha=0.1/seed=42/topk=4/lora_rank=32/lr=4e-05/rounds=100/metrics.json
%/home/ashar236/Projects/Coins/clientmorpher/outputs_old/dolly_dirichlet/split=dirichlet/mode=expert_clustering/clients=8/clusters=4/alpha=0.1/seed=42/topk=4/lora_rank=32/lr=4e-05/rounds=100/metrics.json 
%/home/ashar236/Projects/Coins/clientmorpher/outputs_old/dolly_dirichlet_client_alpha1/split=dirichlet/mode=client_clustering/clients=8/clusters=4/alpha=1.0/seed=42/topk=4/lora_rank=32/lr=4e-05/rounds=100/metrics.json
%/home/ashar236/Projects/Coins/clientmorpher/outputs_old/dolly_dirichlet alpha 1.0 for both client and expert
%local /home/ashar236/Projects/Coins/clientmorpher/results/dolly_dirichlet/split=dirichlet
\begin{table*}[t]
\centering
\caption{Comparison of Local Training, MoE-FedAvg, Client Clustering, and Expert Clustering under different Dirichlet concentration parameters ($\alpha$). Client IDs remain fixed across experiments, while task assignments may differ under different $\alpha$ values. ROUGE-L is reported for each client (higher is better).}
\label{tab:client_results_alpha}
\resizebox{\textwidth}{!}{
\begin{tabular}{c c c c c c c c c}
\toprule
&
\multicolumn{5}{c}{\textbf{$\alpha=0.1$}} &
\multicolumn{3}{c}{\textbf{$\alpha=1.0$}} \\
\cmidrule(lr){2-6}
\cmidrule(lr){7-9}
\textbf{Client} & \textbf{Task} & \textbf{Local} & \textbf{MoE-FedAvg} & \textbf{Client Cluster} & \textbf{Expert Cluster} & \textbf{Task} & \textbf{Client Cluster} & \textbf{Expert Cluster} \\
\midrule

C0 & QA
& 0.4679 & 0.5175 & 0.4732 & 0.4702 & CLS & 0.4833 & 0.4748 \\

C1 & IE & 0.4660 & 0.5215 & 0.5085 & 0.5129 & Summ & 0.5174 & 0.4953 \\

C2 & Summ & 0.3179 & 0.3552 & 0.3849 & 0.3798 & Summ & 0.4253 & 0.4269 \\

C3 & CLS & 0.5383 & 0.5529 & 0.5472 & 0.5327 & CLS & 0.5648 & 0.5835 \\

C4 & QA & 0.4013 & 0.4294 & 0.4767 & 0.4875 & IE & 0.4781 & 0.4766 \\

C5 & IE & 0.4274 & 0.5102 & 0.5323 & 0.5301 & IE & 0.4972 & 0.5176 \\

C6 & IE & 0.4294 & 0.5183 & 0.4940 & 0.4988 & CLS & 0.5549 & 0.5517 \\

C7 & IE & 0.4642 & 0.5627 & 0.5502 & 0.5318 & Summ & 0.4088 & 0.4220 \\

\midrule
\textbf{Average} & -- & 0.4391 & 0.4960 & 0.4959 & 0.4929 & -- & 0.4912 & 0.4936 \\

\midrule
\textbf{Classification} & -- & 0.5383 & \textbf{0.5529} & 0.5472 & 0.5327 & -- & 0.5343 & 0.5367 \\

\textbf{Information Extraction} & -- & 0.4467 & \textbf{0.5282} & 0.5213 & 0.5184 & -- & 0.4876 & 0.4971 \\

\textbf{Closed QA} & -- & 0.4346 & 0.4734 & 0.4750 & \textbf{0.4788 }& -- & -- & -- \\

\textbf{Summarization} & -- & 0.3179 & 0.3552 & \textbf{0.3849} & 0.3798 & -- & 0.4505 & 0.4481 \\

\bottomrule
\end{tabular}}
\end{table*}

\subsection{Federated Setting}

We consider a federated learning setting with $N$ clients, each holding a local dataset $D_i$ drawn from a heterogeneous data distribution. Each client is associated with one or more NLP tasks, including text classification, closed-domain question answering, information extraction, and summarization. To evaluate the proposed method under varying degrees of statistical heterogeneity, we consider two non-IID data partitioning strategies. First, the \emph{Pathological} partition, in which each client receives data exclusively from a single task. Second, the \emph{Dirichlet} partition, where the task distribution of each client is sampled from a Dirichlet distribution with concentration parameter $\alpha$. We evaluate $\alpha=0.1$, which represents highly heterogeneous client distributions, and $\alpha=1.0$, which represents a substantially more balanced non-IID setting.

\subsection{Dataset and Client Partitioning}

The Databricks Dolly-15k instruction-following dataset is used as the source dataset. Dolly-15k consists of instruction-response examples organized into multiple task categories. To simulate a strongly heterogeneous federated learning environment, we construct a pathological non-IID split in which each client receives examples from only one task category based on data distribution. The federated setting contains four tasks:
\begin{itemize}
    \item CLS: classification
    \item QA: closed-domain question answering
    \item IE: information extraction
    \item Summ: summarization
\end{itemize}

This partitioning creates distinct client distributions and allows us to evaluate whether a federated method can avoid negative transfer between task-specific clients. Each example is formatted as an Alpaca-style instruction-following prompt~\cite{alpaca23}. During training, the response is included in the sequence, but the loss is computed only over response tokens by masking the prompt tokens. During evaluation, only the prompt is given to the model, and the generated response is compared with the reference answer.

\subsection{Compared Methods}

We compare the following adaptation strategies: \textit{Local Training}, \textit{MoE-FedAvg}, \textit{Client Clustering}, and \textit{Expert Clustering}.

\subsubsection{Local Training}
In this baseline, each client fine-tunes its own LoRA~\cite{hu2022lora} parameters using only its local data. No model parameters are exchanged or aggregated during training, resulting in zero communication overhead and no cross-client collaboration. Each client learns a fully personalized model tailored to its local data distribution.% This baseline serves as a reference for evaluating the benefits of federated knowledge sharing.

\subsubsection{MoE FedAvg}
MoE-FedAvg implements the standard FedAvg algorithm in the proposed MoE architecture. In each communication round, every client performs local fine-tuning on its private data, after which the server aggregates the trainable non-expert parameters using sample-weighted averaging. The resulting global adaptation state is then broadcast back to all clients for the next training round. The server computes a sample-weighted average over all client updates.

% \begin{equation}
%     \phi^{(r+1)}
%     =
%     \sum_{i=1}^{N}
%     \frac{|\mathcal{D}_i|}
%     {\sum_{j=1}^{N} |\mathcal{D}_j|}
%     \phi_i^{(r+1)} ,
% \end{equation}

% where $\phi_i^{(r+1)}$ is the locally updated trainable state from client $i$. This method produces one shared adaptation state for all clients, regardless of task or distributional differences.
\subsubsection{ClientMorpher-C}
The server first computes an expert-activation profile for each client using the frozen MoE router. Clients are then clustered directly using these profiles, and federated averaging is performed only within each cluster.

\subsubsection{ClientMorpher-E}
This variant first summarizes how experts are used across clients. Experts are clustered by their cross-client usage signatures. Each client is then assigned to the expert cluster where most of its routing mass goes. Federated averaging is again performed within the induced client groups.

\subsection{Baseline Comparison}
\label{subsec:baseline_results}
Table~\ref{tab:baseline_results} compares Local Training, MoE-FedAvg, ClientMorpher-C, and ClientMorpher-E on the four-client pathological split. All federated methods share the same trainable LoRA parameters, which correspond to only 0.0878\% of the full model per communication round. Consequently, all federated methods operate under an identical communication budget, allowing improvements to be attributed solely to more effective client collaboration rather than additional communication. Among the compared methods, ClientMorpher-E achieves the highest average ROUGE-L score (0.4986), outperforming both Local Training (0.4824) and MoE-FedAvg (0.4927). It also achieves the best performance in Classification, Closed Question Answering, and Summarization, while ClientMorpher-C achieves the highest score in Information Extraction. These results suggest that organizing collaboration according to routing signatures enables more effective knowledge sharing than aggregating updates globally, particularly when clients exhibit highly heterogeneous instruction distributions.

\subsection{Task-Wise Results Across Heterogeneity}
\label{subsec:variant_results}
Table~\ref{tab:client_results_alpha} compares ClientMorpher-C and ClientMorpher-E under two heterogeneous data distributions generated using Dirichlet parameters $\alpha=0.1$ and $\alpha=1.0$. Across both settings, the MoE backbone, LoRA configuration, clustering parameters, and training schedule remain identical. Under the highly heterogeneous setting ($\alpha=0.1$), both ClientMorpher variants achieve performance comparable to MoE-FedAvg while substantially improving over Local Training. ClientMorpher-C achieves the best performance on Summarization and remains competitive on Closed Question Answering, whereas ClientMorpher-E performs better on Classification and maintains competitive performance on Information Extraction. As the client distribution becomes more balanced ($\alpha=1.0$), the performance gap between the two variants narrows, with ClientMorpher-E achieving the highest overall average ROUGE-L (0.4936) and stronger performance on Classification and Information Extraction, while ClientMorpher-C remains slightly better on Summarization.
 These observations suggest that the two routing-aware clustering strategies capture complementary aspects of client similarity. Directly clustering clients using activation profiles is particularly effective when routing patterns are highly separated under extreme heterogeneity. In contrast, clustering experts based on shared usage signatures produces more stable collaboration groups as client distributions become less skewed. Finally, because no client is dominated by the Closed Question Answering task under the sampled $\alpha=1.0$ partition, task-specific results for that category are not reported.

\section{Conclusion}
\label{conclusion}
% What was the goal, what we achieved? 
This paper introduces ClientMorpher, a routing-aware, personalized, federated instruction fine-tuning framework for sparse Mixture-of-Experts (MoE) LLMs. We addressed the challenge of collaborative learning under heterogeneous client instruction distributions by leveraging routing signatures as a signal for client data distribution. ClientMorpher forms personalized groups based on routing behavior, enabling clients with similar instruction characteristics to exchange knowledge more effectively. To investigate different perspectives on routing-aware collaboration, we proposed two variants: ClientMorpher-C, which clusters clients directly using expert activation profiles, and ClientMorpher-E, which uses expert clusters via cross-client expert usage signatures before building client groups. Extensive experiments on federated instruction fine-tuning using the Databricks Dolly-15K dataset under both pathological and Dirichlet-based heterogeneous data distributions demonstrate that routing-aware collaboration consistently improves personalized performance over traditional Federated Averaging and local training while maintaining the same communication cost. Our analysis further shows that client-centric and expert-centric clustering capture complementary aspects of task similarity. Direct client clustering is particularly effective under highly skewed client distributions, whereas expert-centric clustering provides more robust collaboration as client heterogeneity decreases. 

\bibliographystyle{IEEEtran}
\bibliography{references}

@inproceedings{ye2024openfedllm,
  title={Openfedllm: Training large language models on decentralized private data via federated learning},
  author={Ye, Rui and Wang, Wenhao and Chai, Jingyi and Li, Dihan and Li, Zexi and Xu, Yinda and Du, Yaxin and Wang, Yanfeng and Chen, Siheng},
  booktitle={Proceedings of the 30th ACM SIGKDD conference on knowledge discovery and data mining},
  pages={6137--6147},
  year={2024}
}

@article{shazeer2017outrageously,
  title={Outrageously large neural networks: The sparsely-gated mixture-of-experts layer},
  author={Shazeer, Noam and Mirhoseini, Azalia and Maziarz, Krzysztof and Davis, Andy and Le, Quoc and Hinton, Geoffrey and Dean, Jeff},
  journal={arXiv preprint arXiv:1701.06538},
  year={2017}
}

@article{mei2024fedmoe,
  title={Fedmoe: Personalized federated learning via heterogeneous mixture of experts},
  author={Mei, Hanzi and Cai, Dongqi and Zhou, Ao and Wang, Shangguang and Xu, Mengwei},
  journal={arXiv preprint arXiv:2408.11304},
  year={2024}
}

@article{zhan2024fedmoe,
  title={FedMoE-DA: Federated mixture of experts via domain aware fine-grained aggregation},
  author={Zhan, Ziwei and Zhao, Wenkuan and Li, Yuanqing and Liu, Weijie and Zhang, Xiaoxi and Tan, Chee Wei and Wu, Chuan and Guo, Deke and Chen, Xu},
  journal={arXiv preprint arXiv:2411.02115},
  year={2024}
}

@article{ghosh2020efficient,
  title={An efficient framework for clustered federated learning},
  author={Ghosh, Avishek and Chung, Jichan and Yin, Dong and Ramchandran, Kannan},
  journal={Advances in neural information processing systems},
  volume={33},
  pages={19586--19597},
  year={2020}
}

@article{hu2022lora,
  title={Lora: Low-rank adaptation of large language models.},
  author={Hu, Edward J and Shen, Yelong and Wallis, Phillip and Allen-Zhu, Zeyuan and Li, Yuanzhi and Wang, Shean and Wang, Liang and Chen, Weizhu and others},
  journal={Iclr},
  volume={1},
  number={2},
  pages={3},
  year={2022}
}

@article{liu2025flex,
  title={FLEx: Personalized Federated Learning for Mixture-of-Experts LLMs via Expert Grafting},
  author={Liu, Fan and Pan, Bikang and Wang, Zhongyi and Yao, Xi and Tang, Xiaoying and Wang, Jingya and Shi, Ye},
  journal={arXiv preprint arXiv:2506.00965},
  year={2025}
}

@article{Yi2024pFedMoEDP,
  title={pFedMoE: Data-Level Personalization with Mixture of Experts for Model-Heterogeneous Personalized Federated Learning},
  author={Liping Yi and Han Yu and Chao Ren and Heng Zhang and Gang Wang and Xiaoguang Liu and Xiaoxiao Li},
  journal={ArXiv},
  year={2024},
  volume={abs/2402.01350},
  url={https://api.semanticscholar.org/CorpusID:267548169}
}

@article{wang2026myth,
  title={The myth of expert specialization in moes: Why routing reflects geometry, not necessarily domain expertise},
  author={Wang, Xi and Hayou, Soufiane and Nalisnick, Eric},
  journal={arXiv preprint arXiv:2604.09780},
  year={2026}
}

@article{fedus2022switch,
  title={Switch transformers: Scaling to trillion parameter models with simple and efficient sparsity},
  author={Fedus, William and Zoph, Barret and Shazeer, Noam},
  journal={Journal of Machine Learning Research},
  volume={23},
  number={120},
  pages={1--39},
  year={2022}
}

@inproceedings{mcmahan2017communication,
  title={Communication-efficient learning of deep networks from decentralized data},
  author={McMahan, Brendan and Moore, Eider and Ramage, Daniel and Hampson, Seth and y Arcas, Blaise Aguera},
  booktitle={Artificial intelligence and statistics},
  pages={1273--1282},
  year={2017},
  organization={Pmlr}
}

@article{kairouz2021advances,
  title={Advances and open problems in federated learning},
  author={Kairouz, Peter and McMahan, H Brendan},
  journal={Foundations and trends in machine learning},
  volume={14},
  number={1-2},
  pages={1--210},
  year={2021},
  publisher={Emerald Publishing Limited}
}

@article{li2020federated,
  title={Federated optimization in heterogeneous networks},
  author={Li, Tian and Sahu, Anit Kumar and Zaheer, Manzil and Sanjabi, Maziar and Talwalkar, Ameet and Smith, Virginia},
  journal={Proceedings of Machine learning and systems},
  volume={2},
  pages={429--450},
  year={2020}
}

@article{tan2022towards,
  title={Towards personalized federated learning},
  author={Tan, Alysa Ziying and Yu, Han and Cui, Lizhen and Yang, Qiang},
  journal={IEEE transactions on neural networks and learning systems},
  volume={34},
  number={12},
  pages={9587--9603},
  year={2022},
  publisher={IEEE}
}

@article{Sattler2019ClusteredFL,
  title={Clustered Federated Learning: Model-Agnostic Distributed Multitask Optimization Under Privacy Constraints},
  author={Felix Sattler and Klaus-Robert M{\"u}ller and Wojciech Samek},
  journal={IEEE Transactions on Neural Networks and Learning Systems},
  year={2019},
  volume={32},
  pages={3710-3722},
  url={https://api.semanticscholar.org/CorpusID:203736521}
}

@online{DatabricksBlog2023DollyV2,
    author    = {Mike Conover and Matt Hayes and Ankit Mathur and Jianwei Xie and Jun Wan and Sam Shah and Ali Ghodsi and Patrick Wendell and Matei Zaharia and Reynold Xin},
    title     = {Free Dolly: Introducing the World's First Truly Open Instruction-Tuned LLM},
    year      = {2023},
    url       = {https://www.databricks.com/blog/2023/04/12/dolly-first-open-commercially-viable-instruction-tuned-llm},
    urldate   = {2023-06-30}
}

@article{avinash2026task,
  title={Task-Conditioned Routing Signatures in Sparse Mixture-of-Experts Transformers},
  author={Avinash, Mynampati Sri Ranganadha},
  journal={arXiv preprint arXiv:2603.11114},
  year={2026}
}

@article{firouzi2026generative,
  title={Generative IoT (GIoT): Advancing IoT with Generative AI and Large Language Models},
  author={Firouzi, Farshad and Ray, Aritra and Farahani, Bahar and Daneshmand, Mahmoud and Song, Jaeseung and Wu, Shaoen and Chakrabarty, Krishnendu},
  journal={Digital Communications and Networks},
  year={2026},
  publisher={Elsevier}
}

@article{jiang2024federated,
  title={Federated clustered multi-domain learning for health monitoring},
  author={Jiang, Shiyi and Li, Yuan and Firouzi, Farshad and Chakrabarty, Krishnendu},
  journal={Scientific reports},
  volume={14},
  number={1},
  pages={903},
  year={2024},
  publisher={Nature Publishing Group UK London}
}

@article{jiang2023low,
  title={Low-overhead clustered federated learning for personalized stress monitoring},
  author={Jiang, Shiyi and Firouzi, Farshad and Chakrabarty, Krishnendu},
  journal={IEEE Internet of Things Journal},
  volume={11},
  number={3},
  pages={4335--4347},
  year={2023},
  publisher={IEEE}
}

@article{zhou2025fedsem,
  title={FedSem: A Resource Allocation Scheme for Federated Learning Assisted Semantic Communication},
  author={Zhou, Xinyu and Li, Yang and Zhao, Jun},
  journal={arXiv preprint arXiv:2503.06058},
  year={2025}
}

@inproceedings{duan2021fedgroup,
  title={Fedgroup: Efficient federated learning via decomposed similarity-based clustering},
  author={Duan, Moming and Liu, Duo and Ji, Xinyuan and Liu, Renping and Liang, Liang and Chen, Xianzhang and Tan, Yujuan},
  booktitle={2021 IEEE Intl Conf on parallel \& distributed processing with applications, big data \& cloud computing, sustainable computing \& communications, social computing \& networking (ISPA/BDCloud/SocialCom/SustainCom)},
  pages={228--237},
  year={2021},
  organization={IEEE}
}

@article{zeng2025stocfl,
  title={StoCFL: A stochastically clustered federated learning framework for Non-IID data with dynamic client participation},
  author={Zeng, Dun and Hu, Xiangjing and Liu, Shiyu and Yu, Yue and Wang, Qifan and Xu, Zenglin},
  journal={Neural Networks},
  volume={187},
  pages={107278},
  year={2025},
  publisher={Elsevier}
}

@article{guo2020pfl,
  title={PFL-MoE: Personalized federated learning based on mixture of experts},
  author={Guo, Binbin and Mei, Yuan and Xiao, Danyang and Wu, Weigang and Yin, Ye and Chang, Hongli},
  journal={arXiv preprint arXiv:2012.15589},
  year={2020}
}

@inproceedings{jiang2025heterogeneous,
  title={Heterogeneous federated learning with scalable server mixture-of-experts},
  author={Jiang, Jingang and Chen, Yanzhao and Liu, Xiangyang and Jiang, Haiqi and Fan, Chenyou},
  booktitle={Proceedings of the Thirty-Fourth International Joint Conference on Artificial Intelligence},
  pages={5480--5488},
  year={2025}
}

@misc{alpaca23,
      title={Facilitating the sharing of electrophysiology data analysis results through in-depth provenance capture},
      author={Cristiano André Köhler and Danylo Ulianych and Sonja Grün and Stefan Decker and Michael Denker},
      year={2023},
      eprint={2311.09672},
      archivePrefix={arXiv},
      primaryClass={q-bio.NC}
}
\end{document}